\documentclass[conference]{IEEEtran}
\IEEEoverridecommandlockouts
\usepackage{cite}
\usepackage{amsmath,amssymb,amsfonts}
\usepackage{algorithmic}
\usepackage{graphicx}
\usepackage{textcomp}
\usepackage{xcolor}
\usepackage{subcaption}
\usepackage{balance}

\usepackage{amsmath}
\usepackage{float}
\usepackage[switch]{lineno}
\newcommand{\ignore}[1]{}

\definecolor{Red}{cmyk}{0,1,1,0}
\definecolor{Blue}{cmyk}{1,1,0,0}
\definecolor{Orange}{rgb}{1,0.5,0}

\def\BibTeX{{\rm B\kern-.05em{\sc i\kern-.025em b}\kern-.08em
    T\kern-.1667em\lower.7ex\hbox{E}\kern-.125emX}}
    
\begin{document}

\title{Effects of the Nonlinearity in Activation Functions on the Performance of Deep Learning Models\\
%{\footnotesize \textsuperscript{*}Note: Sub-titles are not captured in Xplore and should not be used}
\thanks{Funded by the National Science Foundation (NSF), grant no: CNS-1831980}
}

\author{\IEEEauthorblockN{Nalinda Kulathunga}
\IEEEauthorblockA{\textit{Department of Mathematics} \\
\textit{Texas Southern University}\\
Houston, TX, USA. \\
nalinda05kl@gmail.com}

\and

\IEEEauthorblockN{Daniel Vrinceanu}
\IEEEauthorblockA{\textit{Department of Physics} \\
\textit{Texas Southern University}\\
Houston, TX, USA. \\
ORCID: 0000-0002-8820-9073}

\and

\IEEEauthorblockN{Zackary Kinsman}
\IEEEauthorblockA{\textit{Department of Mathematics} \\
\textit{Texas Southern University}\\
Houston, TX, USA. \\
Zackary.Kinsman@tsu.edu}

\and

\IEEEauthorblockN{Nishanth Rajiv Ranasinghe}
\IEEEauthorblockA{\textit{Department of Computer Science} \\
\textit{Prairie View A \& M University}\\
Prairie View, TX, USA. \\
nranasinghe@unm.edu}

\and

\IEEEauthorblockN{Lei Huang}
\IEEEauthorblockA{\textit{Department of Computer Science} \\
\textit{Prairie View A \& M University}\\
Prairie View, TX, USA. \\
lhuang.pvamu@gmail.com}

\and

\IEEEauthorblockN{Yonggao Yang}
\IEEEauthorblockA{\textit{Department of Computer Science} \\
\textit{Prairie View A \& M University}\\
Prairie View, TX, USA. \\
yoyang@pvamu.edu}

\and

\IEEEauthorblockN{Yunjiao Wang}
\IEEEauthorblockA{\textit{Department of Mathematics} \\
\textit{Texas Southern University}\\
Houston, Texas, USA \\
Yunjiao.Wang@tsu.edu}
}

\author{Nalinda Kulathunga\textsuperscript{1}, Nishath Rajiv Ranasinghe\textsuperscript{3}, Daniel Vrinceanu\textsuperscript{2}, \\ Zackary Kinsman\textsuperscript{1}, Lei Huang\textsuperscript{3} and Yunjiao Wang\textsuperscript{1}\\ \textsuperscript{1}\small{Department of Mathematics, Texas Southern University, Houston, TX, USA.}\\ \textsuperscript{2}\small{Department of Physics, Texas Southern University, Houston, TX, USA.} \\ \textsuperscript{3}\small{Department of Computer Science, Prairie View A \& M University, Prairie View, TX, USA.}}

\maketitle

\begin{abstract}
The nonlinearity of activation functions used in deep learning models are crucial for the success of predictive models. There are several commonly used simple nonlinear functions, including Rectified Linear Unit (ReLU) and Leaky-ReLU (L-ReLU). In practice, these functions remarkably enhance the model accuracy. However, there is limited insight into the functionality of these nonlinear activation functions in terms of why certain models perform better than others. Here, we investigate the model performance when using ReLU or L-ReLU as activation functions in different model architectures and data domains. Interestingly, we found that the application of L-ReLU is mostly effective when the number of trainable parameters in a model is relatively small. Furthermore, we found that the image classification models seem to perform well with L-ReLU in fully connected layers, especially when pre-trained models such as the VGG-16 are used for the transfer learning. 
\end{abstract}


\section{Introduction}\label{s1}

The great success of deep learning in applications is based on the clever idea of constructing sufficiently large nonlinear function spaces throughout the composition of layers of linear and simple nonlinear functions (in the name of activation function). Widely used activation functions include the Sigmoidal function, Rectified Linear Unit (ReLU), and its variants Leaky-ReLU (L-ReLU) and Exponential Linear Unit (ELU) \cite{Pedamonti2018,Nwankpa2018,xu2015empirical}. Since it was first introduced by Nair \textit{et al.} \cite{ReLU}, ReLU has become one of the most popular choices of activation function for many deep learning applications \cite{bestReLU}. The ReLU function is defined as; $\text{ReLU}(x) = max(0, x)$, where $x$ stands for the input. The simplistic ReLU function greatly reduces the computational cost  since ReLU$(x) = 0$ when $x<0$. On the other hand, it does cause an information loss in each layer, which could eventually lead to the vanishing gradient problem during the model training \cite{pascanu2012difficulty}. L-ReLU was introduced by Mass \textit{et al.} \cite{LReLU} to overcome the disadvantage of ReLU and it has the form;

\vspace{-4mm}

\begin{equation}\label{E:LReLU}
\text{L-ReLU}(x) = \left\{ \begin{array}{rl}
\alpha x, \text{    }& x<0\\
\\
x ,  \text{    }& x\geq 0
\end{array}\right.
\end{equation}

\noindent
where $\alpha$ is the linearity factor ($0 < \alpha < 1.0$). Note that the L-ReLU with $\alpha = 0$ is the ReLU function and it becomes an identity linear function when $\alpha = 1$. In practice, the value of $\alpha$ is kept closer to zero. However, there is no supporting theory to predict the ideal value of $\alpha$. Instead, it is usually determined through trial-and-error observations. Similarly, selection between ReLU and L-ReLU for a certain network is determined by prior experimental knowledge. Therefore, a study of experimental evidence to enhance our understanding of the behavior of activation functions is necessary in order to minimize the possible speculations.

\par
The nonlinearity of a deep network results from the composition of the layers of  nonlinear functions. Therefore, the impact of different activation functions  on the final performance is likely linked to the model architecture (width and the depth). In addition, as pointed out in \cite{souze2020}, model performance also depends on the  data type (for example; continuous, categorical, calligraphic, or photographic).

\par
\begin{figure}[]
\centerline{\includegraphics[width=55mm]{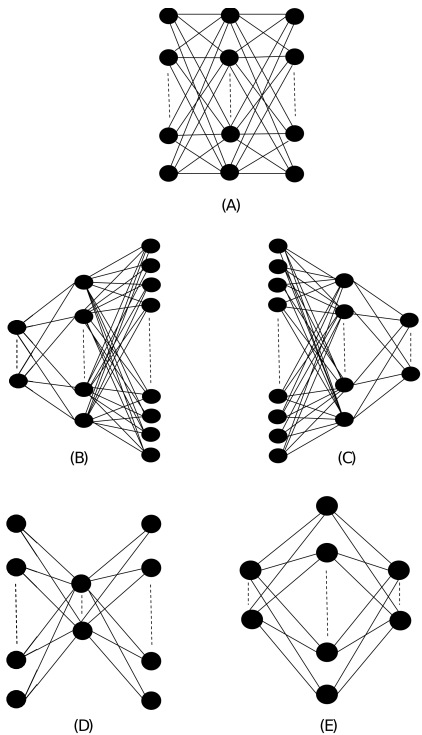}}
\caption{Shapes of different model architectures used in the analysis. For each architecture shape, number of nodes per hidden layer was varied according to the Table~\ref{tab1}, in order to change the model complexity.}
\label{fig1}
\end{figure}

Based on the considerations above, we study how different choices of the linearity factor $\alpha$ impact the model performance by first exploring the effect of the network shapes on validation accuracy, then investigating the effect of $\alpha$ on different network shapes based on the architectures illustrated in Fig.~\ref{fig1} and on different data domains. We found that ReLU performed better most of the time except for when the model did not have sufficient nodes in each layer.

\section{Method}\label{s2}

For convenience, we separate the method into two subsections according to the following two objectives:

\begin{itemize}
\item Objective 1: Investigation of the effect of nonlinearity in the activation function (L-ReLU) on the model accuracy for different model architectures.
\item Objective 2: Investigation of the effects of nonlinearity in the activation function (L-ReLU) on the model performance in the presence of different data domains.
\end{itemize}

\subsection{Method for Objective 1}\label{s2.1}

\subsubsection{Data Set}\label{s2.1.1}

For the Objective 1, the analysis was done using 10,000 images from the MNIST data set \cite{MNIST-data}. The MNIST data set consists of images of hand written digits belongs to ten labeled classes from 0 to 9. Each instance in this data set consists of a $28\times28$ grey-scale pixel image.  

\subsubsection{Model Architecture}\label{s2.1.2} 

For this test, five model architecture shapes (Fig.~\ref{fig1}), each with three hidden layers were used. ReLU or L-ReLU was used as the activation function. For the convenience, we will refer the shape of the network shown in Fig. 1A-1E as Architecture A-E. Architecture A consists of a constant number of nodes per hidden layer while in the other four architectures (Architecture B-E) the number of nodes varies from layer to layer. In Architecture B and C, the ratio between the number of nodes in two consecutive layers is constant while in Architecture D and E, the ratio between the nodes in first two layers is the reciprocal of the ratio between the number of nodes in last two layers. The complexity of each model architecture was varied by changing the number of nodes per hidden layer according to the combinations shown in Table~\ref{tab1}.

\subsubsection{Evaluation Matrix and Optimizer}\label{s2.1.3}

The validation accuracy was used as the evaluation matrix for this part of the analysis. Here, we used the average value of the validation accuracy in last five epochs of the total of 20 epochs used to train the models. 33\% of the total data set was used as the validation data. The stochastic gradient descent \cite{SGD} with a learning rate of 0.1 was used as the optimizer while the categorical cross entropy \cite{cross-entropy} was used as the loss function.

\begin{table}[htbp]
\caption{\centering Different combinations of number of nodes used in three hidden layers of five network architecture shapes shown in in Fig.~\ref{fig1}} 
\begin{center}
\begin{tabular}{c c c c c}
\hline
\textbf{ARC.}&\multicolumn{4}{|c}{\textbf{Number of Nodes in Hidden Layers (width)}} \\
\cline{2-5} 
\textbf{shape} & \multicolumn{1}{|c}{\textbf{width-1}} & \textbf{width-2} & \textbf{width-3} & \textbf{width-4} \\
\hline
A & 8, 8, 8 & 16, 16, 16 & 32, 32, 32 & 64, 64, 64 \\
B & 8, 16, 32 & 16, 32, 64 & 32, 64, 128 & 64, 128, 256 \\
C & 16, 8, 4 & 32, 16, 8 & 64, 32, 16 & 128, 64, 32 \\
D & 16, 8, 16 & 32, 16, 32 & 64, 32, 64 & 128, 64, 128 \\
E & 8, 16, 8 & 16, 32, 16 & 32, 64, 32 & 64, 128, 64 \\
\hline
\end{tabular}
\label{tab1}
\end{center}
\end{table}

\subsubsection{Procedure}\label{s2.1.4}

Each of the model architectures listed in Table~\ref{tab1} was trained using MNIST images and the validation accuracy was recorded for 20 epochs with their uncertainties. In this analysis, for each test, the statistical uncertainties were estimated by calculating the standard error on the mean value for the validation accuracy recorded over 20 iterations. Different models were obtained, a). by changing the number of nodes in a fixed model architecture shape and b). by changing the shape of the model architecture while keeping the total number of parameters fixed ($10^{5}\pm$1\%). This analysis was extended in order to study the effect of nonlinearity in L-ReLU function on the model performance by varying the linearity factor, $\alpha$ from zero (ReLU) to one (identity linear) with 0.1 step size. Resulting validation accuracy was recorded as a function of $\alpha$ with the statistical uncertainties.

\subsection{Method for Objective 2}\label{s2.2}

\subsubsection{Data Sets}\label{s2.2.1}

In this part of the analysis, three data sets were used to test the effects of $\alpha$ on the model performance in the presence of different data domains (continuous and categorical). 
\vspace{-4mm}

\begin{equation}\label{E:cont_data}
f(x) =  \exp(\sum_{n=1}^{16} a_{n}x_{n})
\end{equation}

For the continuous data, we have simulated 10,000 data instances with 16 features using an underlying function; $f(x)$, as shown in Eq.~(\ref{E:cont_data}) with a Gaussian noise of one standard deviation where, $x_{n}$ and $a_{n}$ represent the $n^{th}$ feature and its coefficient, respectively. The MNIST (10,000 images) and the FOOD-11 (5000 images) \cite{FOOD-11-data} data sets were used as the categorical data sets. A validation split of 33\% was used for all three data sets.

\subsubsection{Model Architecture}\label{s2.2.2}

For the continuous data and MNIST data, a network with fully connected layers were used. For the FOOD-11 data set, the model was constructed using bottleneck features, which were extracted using transfer learning \cite{Bottle} from the pre-trained model; VGG16 \cite{VGG16}, followed by a fully connected network. The L-ReLU was used as the activation function in the fully connected layers for all three data domains. In the output layers of these models, the softmax activation function was used for the MNIST and FOOD-11 data while there was no activation function used in the last layer of the model trained using continuous data. The input dimensions for the models with fully connected layers used in the training were $1\times16$, $28\times28$ and $1\times512$ for continuous, MNIST and FOOD-11 data sets, respectively. For all three data sets, the classification model was constructed using the same hidden layer architecture using four hidden layers with 128, 512, 512 and 128 nodes per layer. See the Appendix~\ref{A.D} for more information about the model architectures.

\subsubsection{Evaluation Matrix and Optimizer}\label{s2.2.3}

Mean Squared Error (MSE) was used as the evaluation matrix for the regression analysis. In the classification tasks, for the comparison purposes, categorical cross entropy loss was used as the evaluation matrix instead of the accuracy. Stochastic gradient descent with learning rate of 0.1 was used as the optimizer for all three data sets. For the classification tasks, the categorical cross entropy was used as the loss function while for the regression analysis it was the MSE loss.

\subsubsection{Procedure}\label{s2.2.4}

All three data sets were trained using the same hidden layer architecture in the fully connected network as mentioned above in the Section~\ref{s2.2.2} (and in Appendix~\ref{A.D}) and the average validation loss was calculated. For a certain value of $\alpha$, this process was done 20 times and the statistical uncertainties were calculated using the standard error on the mean value of the recorded validation losses. This procedure was done for $\alpha$ factors starting from zero until one with step size of 0.1. When comparing results form different data sets, the validation loss at each $\alpha$ factor was normalized by the validation loss at $\alpha$ = 0 (ReLU) for each data set. The construction of the deep learning models as well as the model training were done using the deep learning tools (Scikit-learn, TensorFlow-keras and Pytorch) available in \cite{sklearn, keras_cite, Pytorch_cite}. Refer to our GitHub repository\footnote{GitHub: https://github.com/nalinda05kl/nonlinear\_activations} for detailed analysis codes used in this study. The computations were conducted mainly using the Pittsburgh Super-computing Center (PSC) \cite{Nystrom:2015:BUF:2792745.2792775}.

\section{Results and Discussion}\label{s3}

Here, we present the results about the effects of network architecture on the model performance due to the variations in number of parameters, shape of the network, and the nonlinearity. We also studied the effects of different data domains on the model performance. Variations in the aforementioned hyper-parameters were done over a wide range and the results are discussed in this section. However, we will only show the results needed to convey the main messages in this section and the additional figures and tables are included in Appendix~\ref{A.A}-\ref{A.D}.

\subsection{Effects of Number of Parameters}\label{s3.1}

\begin{figure}[ht]
\centerline{\includegraphics[width=80mm]{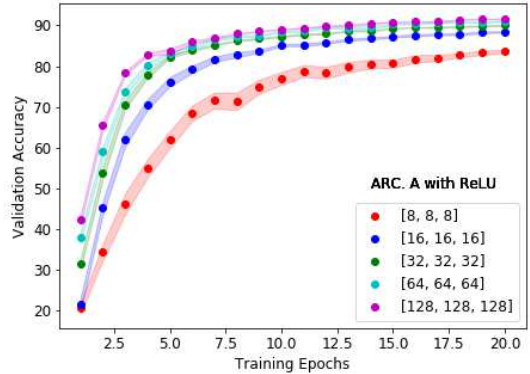}}
\caption{Validation accuracy of the model Architecture A (ARC. A) for 20 training epochs using 10,000 images from MNIST data set for different number of node combinations. ReLU was used as the activation function.}
\label{fig2}
\end{figure}

First, we tested the dependence of the number of parameters on the model performance using ReLU for MNIST data set. Results for the model Architecture A is shown in Fig.~\ref{fig2}. As expected, the model performance improved when the number of parameters was increased. Similar to model Architecture A, all of the other model architectures showed the improvement of accuracy when the number of parameters were increased (Appendix~\ref{A.A}). This shows that a greater availability of trainable parameters helps the learning process regardless of the the shape of the model architecture. However, the determination of the upper bound for the number of parameters should be done by imposing an early stopping criteria \cite{EARLYSTOP} for better generalization of the model.

\begin{figure}[h]
\centerline{\includegraphics[width=80mm]{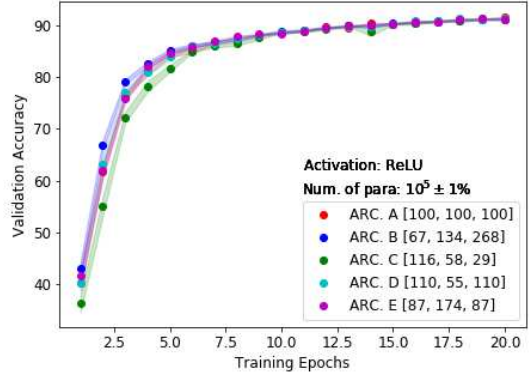}}
\caption{Validation accuracy during 20 training epochs for neural network models with a fixed number of parameters ($10^{5}\pm1\%$) and using 10,000 images from MNIST data set for five different model architectures illustrated in Fig.~\ref{fig1}}
\label{fig3sfig1}
\end{figure}

\begin{figure}[h]
\centerline{\includegraphics[width=80mm]{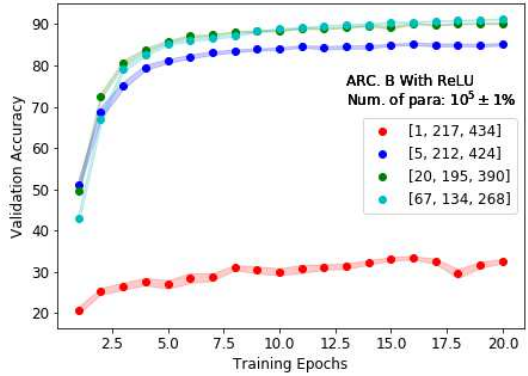}}
\caption{Validation accuracy during 20 training epochs for neural network models with a fixed number of parameters ($10^{5}\pm1\%$) and using 10,000 images from MNIST data set for the model Architecture B (ARC. B) tested for different $1^{st}$ layer widths. ReLU was used as the activation function.}
\label{fig3sfig2}
\end{figure}

\subsection{Effects of Shape of the Network Architecture}\label{s3.2}

Next, we examined model performance for different shapes of model architectures. We first explored the case with the total number of parameters fixed to be $10^{5}$ $(\pm1\%)$ and with ReLU as the activation function. The result is shown in Fig.~\ref{fig3sfig1}, which indicated that all five types of network shapes ended in almost the same result. That is, there was no significant impact on the model performance due to the model architecture. The question now becomes whether these results can be generalized. Next, we studied one type of architecture shape (Architecture B) and varied the number of nodes at the first layer while keeping the total number of parameters fixed as before. As shown in Fig~\ref{fig3sfig2}, the network model with the first layer consisting of one node (in red) has very low performance: $\sim$30\% validation accuracy. The performance was much improved when the number of the nodes at the first layer increased to five nodes ($\sim$80\% validation accuracy). When the number of nodes was set to 20 at the first layer, the validation accuracy was increased to $\sim$90\%. However, further increase in the number of nodes at the first layer did not show improvement. 

The above observation can be understood as follows. When the first layer consists of only one node or insufficient number of nodes, only a small subset of features  were passed through to the next layers, which led to the loss of some essential information needed to make correct predictions.  The same reasoning can be applied to other layers and network architectures. That is, each layer has to have sufficient number of nodes in order to avoid losing critical information. On the other hand, when all layers have sufficient number of nodes, the shape seems to be not crucial. As for how many is sufficient, that depends on the properties of data. As we see from Fig.~\ref{fig2} and \ref{fig3sfig1}, when each layer has sufficiently large number of nodes,  the total number of parameters plays the main role.

\subsection{Effect of Network-Nonlinearity on Model Performance}\label{s3.3}

\begin{figure}[h]
\centerline{\includegraphics[width=75mm]{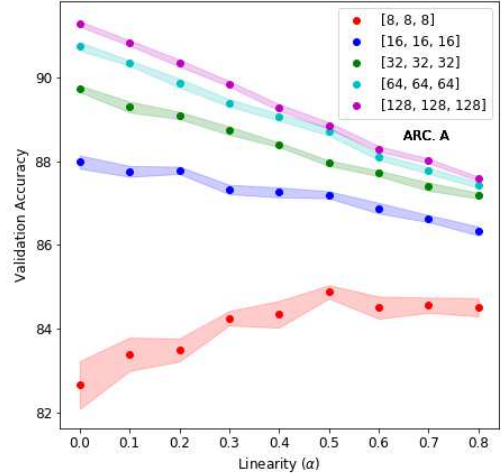}}
\caption{The average validation accuracy as a function of linearity factor ($\alpha$) for neural network models trained using 10,000 images from MNIST data set for the model Architecture A (ARC. A) for different number of parameters.}
\label{fig4sfig1}
\end{figure}

\begin{figure}[h]
\centerline{\includegraphics[width=75mm]{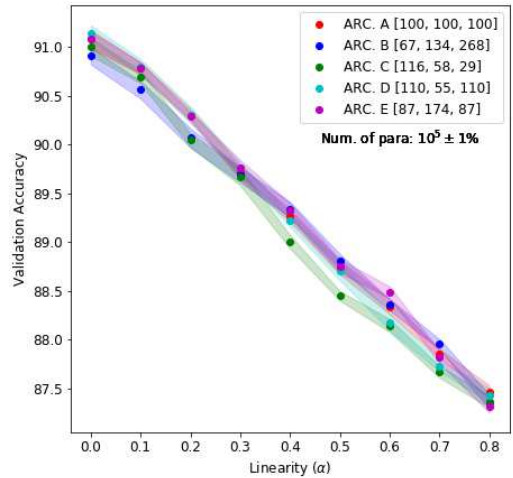}}
\caption{The average validation accuracy as a function of linearity factor ($\alpha$) for neural network models trained using 10,000 images from MNIST data set for different model architectures (ARC. A-E) with fixed number of parameters ($10^{5}$ $\pm$1\%).}
\label{fig4sfig2}
\end{figure}

Next, we explored the effect of the linearity factor, $\alpha$ on model performance based on the five architectures in Fig.~\ref{fig1} with different total number of parameters. Fig.~\ref{fig4sfig1} shows the average validation accuracy for Architecture A as a function of $\alpha$ with the statistical uncertainties measured by standard deviation. As we have seen in the models with ReLU in Section~\ref{s3.1},  the model accuracy increased as the number of parameters increased for all values of $\alpha$. Similar observations can be seen for the other four Architectures B, C, D and E (Appendix~\ref{A.B}). One interesting observation in Fig.~\ref{fig4sfig1} (and in Appendix~\ref{A.B}) is that the L-ReLU was able to increase the validation accuracy only when the number of parameters were relatively small, regardless of the shape of the model architecture. Based on these observations, in Table~\ref{tab2} we show the best performing activation function (ReLU:R or L-ReLU:LR) for each model architecture shape tested with different node combinations. A possible reason for L-ReLU being a better alternative to ReLU for  small networks could be that ReLU has suppress some essential information that were preserved by L-ReLU (in a regularized fashion). However, when the network becomes wider, the increase in the number of parameters enlarges feature representation space so as to efficiently capture more essential features with ReLU than with L-ReLU.

\begin{table}[t]
\begin{center}
\caption{\centering Best activation function (AF.); ReLU (R) or L-ReLU (LR) for different model architectures (AR.). Note: A-8 means the Architecture A with 8 nodes in the first layer of three-layer neural network (see Table~\ref{tab1} for details)} 
\begin{tabular}{c c c c c c c c}
\hline
\multicolumn{2}{c}\centering\textbf{width-1}\hspace{0.4cm} & 
\multicolumn{2}{c}\centering\textbf{width-2\hspace{0.4cm}} & 
\multicolumn{2}{c}\centering\textbf{width-3\hspace{0.4cm}} & 
\multicolumn{2}{c}\centering\textbf{width-4\hspace{0.4cm}} \\
\hline
\textbf{AR.} & \textbf{AF.} & \textbf{AR.} & \textbf{AF.} & \textbf{AR.} & \textbf{AF.} & \textbf{AR.} & \textbf{AF.}\\
\hline
A-8 & LR & A-16 & R & A-32 & R & A-64 & R\\
B-8 & LR & B-16 & LR & B-32 & R & B-64 & R\\
C-16 & LR & C-32 & LR & C-64 & R & C-128 & R\\
D-16 & LR & D-32 & R & D-64 & R & D-128 & R\\
E-8 & LR & E-16 & LR & E-32 & R & E-64 & R\\
\hline
\end{tabular}
\label{tab2}
\end{center}
\end{table}

\textbf{Remark:} Note that L-ReLU enables the preservation of features with negative values throughout the layers. To preserve such features in the presence of ReLU,  we could initialize the bias to certain positive values instead of zero or near zero as usually done in neural network models. A quick test showed that at least for MNIST data, this strategy helps (Appendix~\ref{A.C}), where using the ReLU with the bias initialized to 0.5 in each layer achieved similar performance as using L-ReLU with nonlinearity = 0.8 (Validation accuracy for ReLU with bias = 0.5 is 85.6\% and for L-ReLU with $\alpha$ = 0.8, it is 84.5\% ).

The effect of the shape of the network architecture on the model performance as a function of linearity factor, $\alpha$ was then tested as shown in Fig.~\ref{fig4sfig2}. Here, the number of parameters were kept approximately constant ($10^{5}$ $\pm$1\%) while changing the shape of the model architecture (from A to E). Interestingly, as shown in Fig.~\ref{fig4sfig2}, we observe that there is no effect of the shape of the model architecture on the accuracy for ReLU and for all values of $\alpha$ in L-ReLU. This observation reveals that the model performance is not highly correlated with the shape of the network architecture in comparison to the correlation of the model performance with the number of parameters. It seems that the most important factor for model accuracy is the number of parameters rather than the shape of the model architecture or the nonlinearity in the activation function.

\subsection{Effects of Nonlinearity for Different Data Domains}\label{s3.4}

\begin{figure}[ht]
\centerline{\includegraphics[width=85mm]{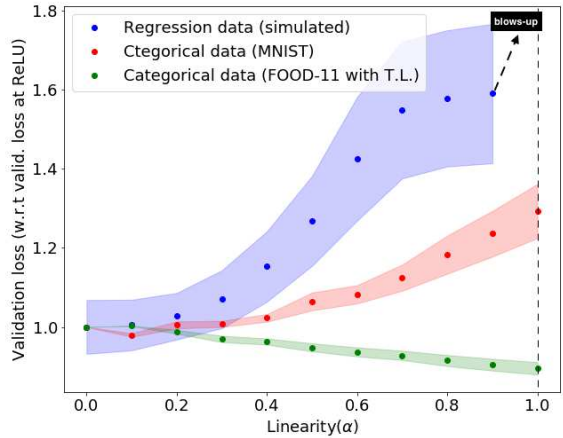}}
\caption{Relative validation loss evaluated as a function of linearity factor ($\alpha$) using a fixed hidden layer architecture for three different data sets: simulated-continuous data (in blue), MNIST data set (in red) and FOOD-11 data set (in green) with transfer learning (T.L.).}
\label{fig5}
\end{figure}

Finally, In Fig.~\ref{fig5}, we show the trends in validation loss as a function of linearity factor, $\alpha$ in L-ReLU using three different data sets; continuous (simulated), MNIST and FOOD-11. In this test, the shape of the model architecture and the number of nodes in each hidden layer of the fully connected network were set to a fixed sequence as explained in the method section (also see Appendix~\ref{A.D}). Due to the use of MSE loss function, the validation loss for the model trained on continuous data was significantly larger compared to other two classification tasks (MNIST, FOOD-11). Therefore, in Fig.~\ref{fig5}, for each data set, the validation loss is normalized by the validation loss evaluated at $\alpha$ = 0, in order to make them comparable.

According to Fig.~\ref{fig5}, for the continuous data, the validation loss increases significantly as a function of $\alpha$. The uncertainty associated with the validation loss is also significantly larger for the continuous data since the MSE was used as the loss function. Note that the training loss for the continuous data reaches an extremely large value at $\alpha = 1$ (in the order of $10^{4}$). This demonstrates the need for ReLU when training data consist of features nonlinearly related to the target variable (see Eq.~(\ref{E:cont_data})). The dependence of validation loss on $\alpha$ for the MNIST data is much weaker compared to the continuous data and it does not blow-up even at $\alpha = 1$. This may be due to the nonlinearity in the softmax activation function applied at the last layer of the model.  

The ReLU function adds strong nonlinearity to the network in comparison to the L-ReLU and enhances the independent learning in each hidden layer. This could be one possible reason for why the validation loss at $\alpha$ = 0 happens to be the best value for both the continuous and MNIST data sets. On the other hand, the dependence of the validation loss on the $\alpha$ factor for the FOOD-11 data shows a contrasting behavior compared to the models used to fit the continuous and MNIST data sets. The validation loss for FOOD-11 data reaches a maximum at $\alpha$ = 0 and decreases as a function of $\alpha$. Note that the model for the FOOD-11 data was trained using the bottleneck features obtained from the VGG-16 \cite{VGG16} pre-trained model (see Appendix~\ref{A.D}, Table~\ref{tab5}). Due to the transfer learning, we can assume that, most of the important information is already extracted into the bottleneck layer. In addition, the nonlinearity is already applied to VGG-16 through ReLU activation function and max-pooling operations. Therefore, the use of a strong nonlinear activation function (ReLU) to classify the bottleneck layer seems to be not improving the accuracy. Instead, there seems to be a loss of important information due to the application of ReLU for the FOOD-11 data.

The above results suggest that the amount of information passed to the next layer by the activation function has a significant effect on the model performance. We find that, it is worth quantifying the level of information flow in the network and analyzing how it is affected by the nonlinearity of activation functions. Therefore, in our future studies, we expect to use entropy as a quantitative measurement for the information flow in a neural network to better understand the effects of the nonlinearity in deep learning.

\section{Conclusions}\label{s4}

We have extensively analysed the effects of the number of parameters, shape of the model architecture and nonlinearity on the performance of deep learning models. We observe, for a given network architecture, validation accuracy increases (up to $\sim$90\%) as a function of the number of parameters regardless of the shape of the model architecture according to the tests we conducted using the MNIST data with the ReLU and L-ReLU activation functions. However, the model shape matters when the first layer dimensions are not enough to capture the high dimensional feature spaces such as those in image classification problems. Interestingly, we found that the L-ReLU function is more effective than ReLU only when the number of parameters is relatively small, according to the tests conducted using MNIST data (all observations are summarized in Table~\ref{tab2}). 

Finally, we looked into different data domains (continuous, categorical with or without transfer learning) as inputs to the networks and their effects on the model performance under different nonlinearities in the network. For the transfer learning, we used a pre-trained model (VGG-16) to extract bottleneck features from image data and classified it using a network with four fully connected hidden layers under different nonlinearities. We found, the use of L-ReLU in the final fully connected layers shows better accuracy than the use of a strong nonlinear function like ReLU for the FOOD-11 data set. In addition to that, we showed the importance of applying nonlinear activation functions in the network to avoid the under-fitting for data consisting features that are nonlinearly related to the targets using simulated-continuous data.

\section*{Acknowledgment}\label{s5}

This work is funded by the National Science Foundation (NSF), grant no: CNS-1831980 and HRD-1800406. Initial computational work in this study was conducted using computational resources at high performance computing centers at Texas Southern University and Prairie View A\&M University. This work used the Extreme Science and Engineering Discovery Environment (XSEDE), which is supported by National Science Foundation grant number ACI-1548562. Specifically, it used the Bridges system, which is supported by NSF award number ACI-1445606, at the Pittsburgh Supercomputing Center (PSC). Authors thank Alice Xu for her contribution in English editing.

%%%%%%%%%%%%%%%%%%%%%%%%% BIBLIOGRAPHY %%%%%%%%%%%%%%%%%%%%%%

\bibliographystyle{IEEEtran}
\bibliography{main}

%%%%%%%%%%%%%%%%%%%%%%%%%%% APPENDIX %%%%%%%%%%%%%%%%%%%%%%%%

\onecolumn

\begin{appendices}
\section{Learning curves for model Architectures B-E}\label{A.A}

\renewcommand{\thefigure}{A.1}
\begin{figure}[h]
\begin{subfigure}{90mm}
  \centering
  \includegraphics[width=80mm]{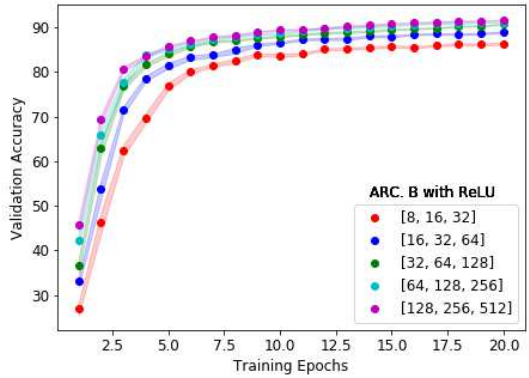}
  \caption{}
  \label{figAB:sfig1}
\end{subfigure}%
\begin{subfigure}{90mm}
  \centering
  \includegraphics[width=80mm]{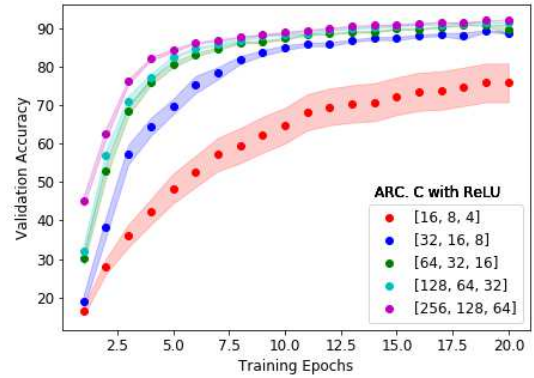}
  \caption{}
  \label{figAB:sfig2}
\end{subfigure}
\begin{subfigure}{90mm}
  \centering
  \includegraphics[width=80mm]{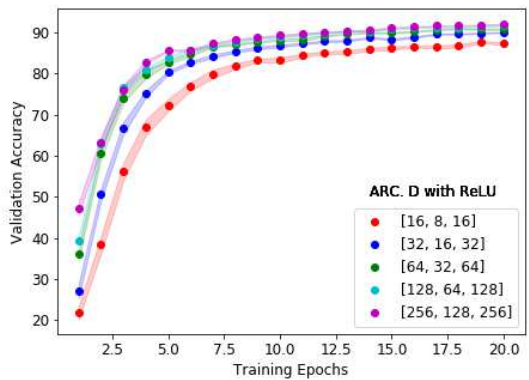}
  \caption{}
  \label{figAB:sfig3}
\end{subfigure}%
\begin{subfigure}{90mm}
  \centering
  \includegraphics[width=80mm]{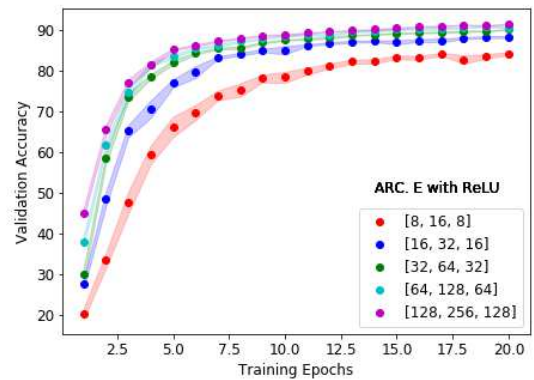}
  \caption{}
  \label{figAB:sfig4}
\end{subfigure}
\caption{Validation accuracy of the model architecture; (a) B, (b) C, (c) D and (d) E for 20 training epochs using 10,000 images from MNIST data set for different number of node combinations. ReLU was used as the activation function.}
\label{figAB:figAB}
\end{figure}

\newpage

\section{Accuracy vs. linearity factor ($\alpha$)}\label{A.B}

\renewcommand{\thefigure}{B.1}
\begin{figure}[h]
\begin{subfigure}{60mm}
  \centering
  \includegraphics[width=60mm]{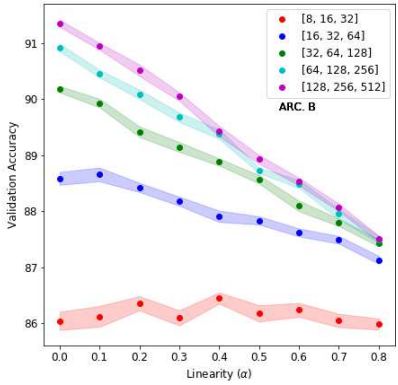}
  \caption{}
  \label{figAC:sfig1}
\end{subfigure}%
\begin{subfigure}{60mm}
  \centering
  \includegraphics[width=60mm]{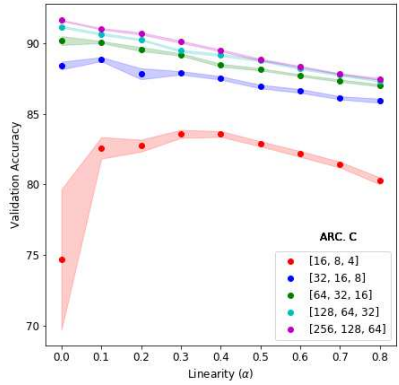}
  \caption{}
  \label{figAC:sfig2}
\end{subfigure}%
\begin{subfigure}{60mm}
  \centering
  \includegraphics[width=60mm]{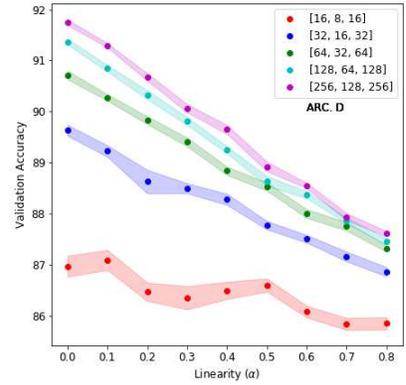}
  \caption{}
  \label{figAC:sfig3}
\end{subfigure}
\begin{subfigure}{60mm}
  \centering
  \includegraphics[width=60mm]{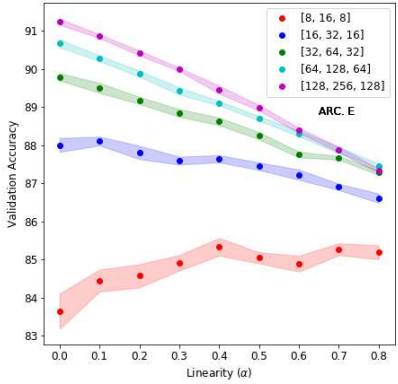}
  \caption{}
  \label{figAC:sfig4}
\end{subfigure}%
\begin{subfigure}{60mm}
  \centering
  \includegraphics[width=60mm]{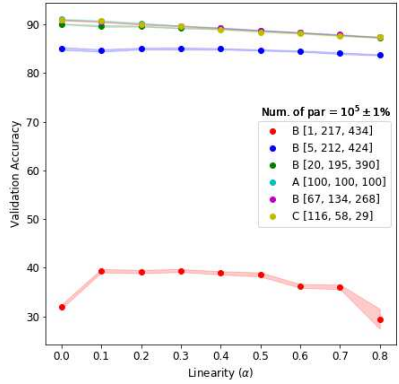}
  \caption{}
  \label{figAC:sfig5}
\end{subfigure}%

\caption{The average validation accuracy as a function of linearity factor ($\alpha$) for the model architecture; (a) B, (b) C, (c) D and (d) E for different number of parameters using 10,000 images from MNIST data set. (e) The average validation accuracy as a function of linearity factor ($\alpha$) for the model architectures A, B and C for different number of parameters using 10,000 images from MNIST data set. Shape B was tested for different widths of the first hidden layer while maintaining the number of parameters fixed ($10^{5}\pm1\%$).}
\label{figAC:figAC}
\end{figure}

\newpage

\section{Comparison of ReLU and L-ReLU with respect to the initial bias.}\label{A.C}

\renewcommand{\thefigure}{C.1}
\begin{figure}[h]
\begin{subfigure}{90mm}
  \centering
  \includegraphics[width=90mm]{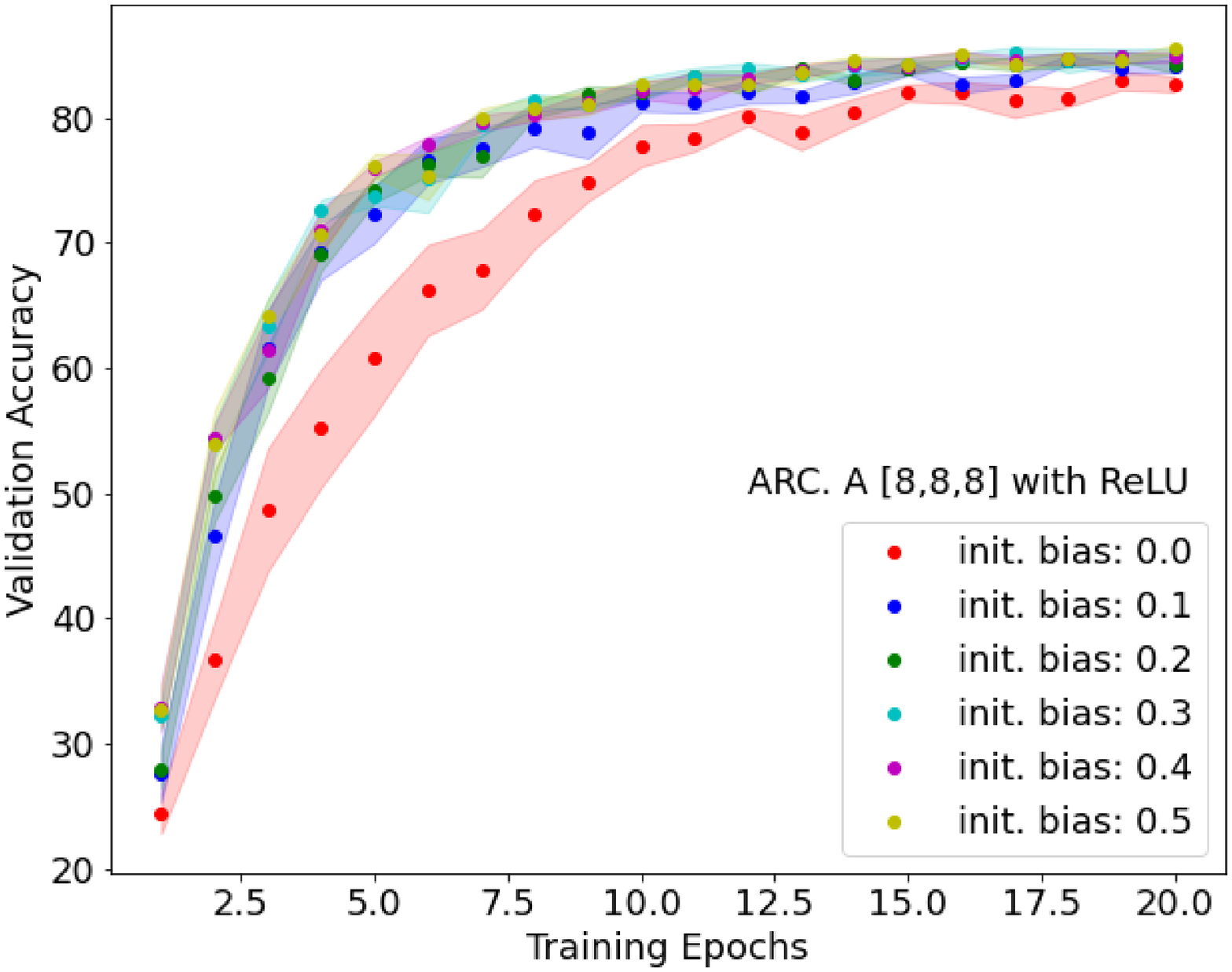}
  \caption{}
  \label{figAC2:sfig1}
\end{subfigure}%
\begin{subfigure}{90mm}
  \centering
  \includegraphics[width=90mm]{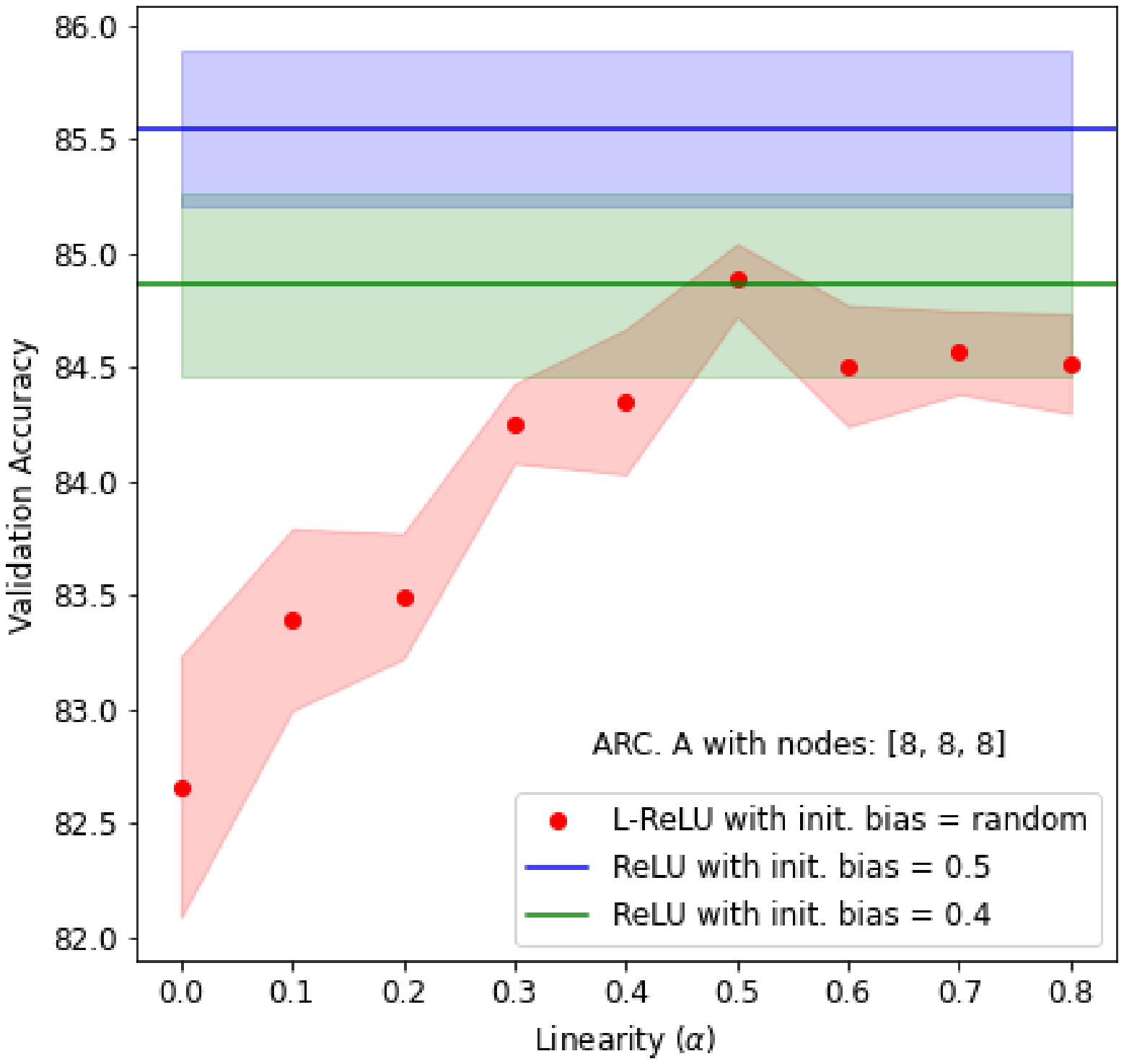}
  \caption{}
  \label{figAC2:sfig2}
\end{subfigure}
\caption{(a) The validation accuracy for the model Architecture A with node combinations: (8, 8, 8) in three hidden layers trained using 10000 MNIST images. Different colors show the learning curves for models initialized using different bias values in the range 0.0 to 0.5 with a step size of 0.1. ReLU was used as the activation. (b) The red curve shows the average validation accuracy as a function of linearity factor ($\alpha$) for the model Architecture A with nodes: (8, 8, 8) in three hidden layers using 10,000 images from MNIST data set. Blue and Green lines show the validation accuracy after 20 training epochs based on same model architecture shape but trained using ReLU with different initial bias values. The color bands show the estimated uncertainties.}
\label{figAC2:figAC2}
\end{figure}

\newpage

\section{Model architectures used in Section~\ref{s3.4}}\label{A.D}

\renewcommand{\thetable}{D.1}
\begin{table}[h]
\caption{\centering Model Architecture for the regression model for continuous data (simulated as in Eq.~(\ref{E:cont_data})) used in the test for the dependence of model performance on data domain for different linearity factors ($\alpha$)} 
\begin{center}
\begin{tabular}{c c c}
\hline
LAYER & SHAPE (in, out) & NUM. of PAR. \\
\hline
Layer-1 & (16, 128) & 2176 \\
Layer-2 & (128, 512) & 66048 \\
Layer-3 & (512, 512) & 262656 \\
Layer-4 & (512, 128) & 65664 \\
Output & (128, 1) & 129 \\
\hline
\end{tabular}
\label{tab3}
\end{center}
\end{table}

\renewcommand{\thetable}{D.2}
\begin{table}[h]
\caption{\centering Model Architecture for MNIST classification used in the test for the dependence of model performance on data domain for different linearity factors ($\alpha$)} 
\begin{center}
\begin{tabular}{c c c}
\hline
LAYER & SHAPE (in, out) & NUM. of PAR. \\
\hline
Layer-1 & (784, 128) & 100480 \\
Layer-2 & (128, 512) & 66048 \\
Layer-3 & (512, 512) & 262656 \\
Layer-4 & (512, 128) & 65664 \\
Output & (128, 10) & 1290 \\
\hline
\end{tabular}
\label{tab4}
\end{center}
\end{table}

\renewcommand{\thetable}{D.3}
\begin{table}[h]
\caption{\centering Model Architecture for FOOD-11 classification used in the test for the dependence of model performance on data domain for different linearity factors ($\alpha$). Note that the input for this model is the bottleneck features extracted from VGG-16 followed by \textit{GlobalAveragePooling2D} operation. The resulting input shape is: 1 $\times$ 512} 
\begin{center}
\begin{tabular}{c c c}
\hline
LAYER & SHAPE (in, out) & NUM. of PAR. \\
\hline
Layer-1 & (512, 128) & 65664 \\
Layer-2 & (128, 512) & 66048 \\
Layer-3 & (512, 512) & 262656 \\
Layer-4 & (512, 128) & 65664 \\
Output & (128, 11) & 1419 \\
\hline
\end{tabular}
\label{tab5}
\end{center}
\end{table}

\end{appendices}

\end{document}